\documentclass[journal, table, doublecolumn]{IEEEtran}
\usepackage[nolist]{acronym}
\usepackage[utf8]{inputenc}
\usepackage[english]{babel}
\usepackage[pdftex]{graphicx}
\usepackage{dblfloatfix}
\usepackage{overpic}
\usepackage{float}
\usepackage{microtype}

\usepackage{enumitem}
\usepackage{url}

\usepackage[binary-units=true, separate-uncertainty]{siunitx}
\usepackage{tabularx, booktabs}

\usepackage{acronym}
\usepackage{todonotes}
\usepackage{amsmath}
\usepackage{amsfonts}
\usepackage{lipsum}

\usepackage[hidelinks]{hyperref}
\usepackage[english, nameinlink, capitalise]{cleveref}
\usepackage{multirow}

\usepackage{scalerel}
\usepackage{tikz}
\usetikzlibrary{svg.path}
\definecolor{orcidlogocol}{HTML}{A6CE39}
\tikzset{
  orcidlogo/.pic={
    \fill[orcidlogocol] svg{M256,128c0,70.7-57.3,128-128,128C57.3,256,0,198.7,0,128C0,57.3,57.3,0,128,0C198.7,0,256,57.3,256,128z};
    \fill[white] svg{M86.3,186.2H70.9V79.1h15.4v48.4V186.2z}
                 svg{M108.9,79.1h41.6c39.6,0,57,28.3,57,53.6c0,27.5-21.5,53.6-56.8,53.6h-41.8V79.1z M124.3,172.4h24.5c34.9,0,42.9-26.5,42.9-39.7c0-21.5-13.7-39.7-43.7-39.7h-23.7V172.4z}
                 svg{M88.7,56.8c0,5.5-4.5,10.1-10.1,10.1c-5.6,0-10.1-4.6-10.1-10.1c0-5.6,4.5-10.1,10.1-10.1C84.2,46.7,88.7,51.3,88.7,56.8z};
  }
}
\newcommand{\orcidicon}[1]{\href{https://orcid.org/#1}{\mbox{\scalerel*{
\begin{tikzpicture}[yscale=-1,transform shape]
\pic{orcidlogo};
\end{tikzpicture}
}{|}}}}

\usepackage[backend=bibtex, bibencoding=utf-8, style=ieee, sorting=none, doi=true, url=false, isbn=false, style=numeric-comp, maxbibnames=2]{biblatex}

\usepackage{geometry}
\usepackage[caption=false,font=normalsize,labelfont=sf,textfont=sf]{subfig}
\usepackage{textcomp}

\usepackage{url}
\usepackage{verbatim}
\usepackage{graphicx}
\def\BibTeX{{\rm B\kern-.05em{\sc i\kern-.025em b}\kern-.08em
    T\kern-.1667em\lower.7ex\hbox{E}\kern-.125emX}}
\usepackage{balance}

\uspunctuation
\begin{acronym}
\acro{AF}{additional features}
\acro{TDC}{time-to-digital converter}
\acro{ASIC}{application-specific integrated circuit}
\acro{TOF}{time-of-flight}
\acro{SNR}{signal-to-noise ratio}
\acro{COG}{center of gravity}
\acro{CTR}{coincidence time resolution}
\acro{cal}{detector under calibration}
\acro{coinc}{coincidence detector}
\acro{DOI}{depth of interaction}
\acro{GTB}[GBDT]{gradient boosted decision trees}
\acro{IR}{isotonic regression}
\acro{kNN}{$k$ nearest neighbors}
\acro{LSF}{line spread function}
\acro{ML}{Machine Learning}
\acro{MAE}{mean absolute error}
\acro{LYSO}{Lutetium–yttrium oxyorthosilicate}
\acro{SR}{spatial resolution}
\acro{SD}{slab detector}
\acro{OD}{one-to-one detector}
\acro{SPAD}{single photon avalanche diode}
\acro{PCA}{principal component analysis}
\acro{PDPC}{Philips Digital Photon Counting}
\acro{PSF}{point spread function}
\acro{RR}{retroreflector}
\acro{PET}{positron emission tomography}
\acro{FWHM}{full width at half maximum}
\acro{ESR}{enhanced specular reflector}
\acro{MLITE}{maximum likelihood interaction time estimation}
\acro{TDL}{tapped delay line}
\acro{LOR}{line-of-response}
\acro{BGO}{Bismuth germanate}
\acro{SiPM}{silicon photomultiplier}
\acro{CNN}{convolutional neural network}
\acro{LED}{leading edge discriminator}
\acro{HF}{high-frequency}
\acro{CNN}{convolutional neural network}
\acro{CT}{computed tomography}
\acro{MRI}{magnetic resonance imaging}
\acro{ToT}{time-over-threshold}
\acro{AI}{Artificial Intelligence}
\end{acronym}

\newcommand{\scBroad}{8.5}
\newcommand{\subFigX}{0}
\newcommand{\subFigY}{80}
\PassOptionsToPackage{table,xcdraw}{xcolor}

\begin{document}
\title{Improving the Timing Resolution of Positron Emission Tomography Detectors using Boosted Learning - A Residual Physics Approach}
\author{Stephan Naunheim$^{*}$\orcidicon{0000-0003-0306-7641}, Yannick Kuhl$^{*}$\orcidicon{0000-0002-4548-0111}, \mbox{David Schug$^{*}$\orcidicon{0000-0002-5154-8303}}, Volkmar Schulz$^{**\dagger}$\orcidicon{0000-0003-1341-9356}, \mbox{Florian Mueller$^{*\dagger}$\orcidicon{0000-0002-9496-4710}}
\thanks{S.Naunheim, Y.Kuhl, and F.Mueller are with the Department of Physics of Molecular Imaging Systems, Institute for Experimental Molecular Imaging, RWTH Aachen University (mail: stephan.naunheim@pmi.rwth-aachen.de). D.Schug is with the Department of Physics of Molecular Imaging Systems, Institute for Experimental Molecular Imaging, RWTH Aachen University, and the Hyperion Hybrid Imaging Systems GmbH, Aachen, Germany. V.Schulz is with the Department of Physics of Molecular Imaging Systems, Institute for Experimental Molecular Imaging, RWTH Aachen University, the Hyperion Hybrid Imaging Systems GmbH, the Fraunhofer MEVIS, Institute for Digital Medicine, and the Physics Institute III B, RWTH Aachen University.\\
$^{*}$IEEE Member, $^{**}$IEEE Senior Member, $^{\dagger}$ V.Schulz and F.Mueller share the last autorship.}}


\markboth{accepted in IEEE TNNLS, DOI: \href{https://doi.org/10.1109/TNNLS.2023.3323131}{10.1109/TNNLS.2023.3323131}\\Please cite the accepted version}%
{How to Use the IEEEtran \LaTeX \ Templates}

\maketitle

\begin{abstract}
\ac{AI} is entering medical imaging, mainly enhancing image reconstruction. Nevertheless, improvements throughout the entire processing, from signal detection to computation, potentially offer significant benefits. This work presents a novel and versatile approach to detector optimization using machine learning and residual physics. We apply the concept to \ac{PET}, intending to improve the \ac{CTR}.\newline
\ac{PET} visualizes metabolic processes in the body by detecting photons with scintillation detectors. Improved \ac{CTR} performance offers the advantage of reducing radioactive dose exposure for patients. Modern \ac{PET} detectors with sophisticated concepts and read-out topologies represent complex physical and electronic systems requiring dedicated calibration techniques. Traditional methods primarily depend on analytical formulations successfully describing the main detector characteristics. However, when accounting for higher-order effects, additional complexities arise matching theoretical models to experimental reality.\newline
Our work addresses this challenge by combining traditional calibration with \ac{AI} and residual physics, presenting a highly promising approach.
We present a residual physics-based strategy using gradient tree boosting and physics-guided data generation. The explainable \ac{AI} framework SHAP was used to identify known physical effects with learned patterns. In addition, the models were tested against basic physical laws. We were able to improve the \ac{CTR} significantly (more than \qty{20}{\percent}) for clinically relevant detectors of \qty{19}{\milli \metre} height, reaching \acp{CTR} of \qty{185}{\pico \second} (\qtyrange{450}{550}{\kilo \eV}).
\end{abstract}


\acresetall
\section{Introduction}

\begin{figure}[b]
\centering
\includegraphics[width=8 cm]{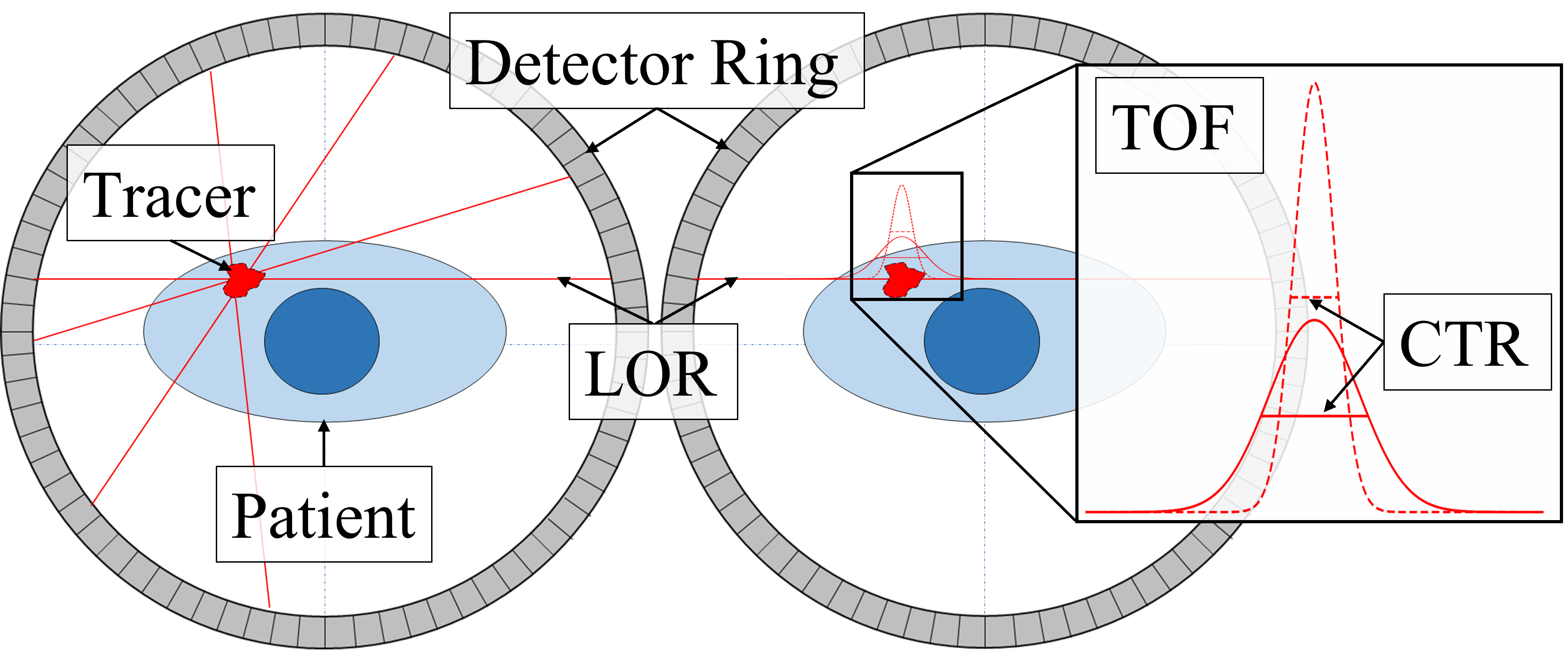}
\caption{Principle of \ac{PET}. A radioactive tracer (red blob) is administered to a patient (bluish). By detecting the created back-to-back $\gamma$-photons, defining many \acp{LOR} (red lines), with the detector ring (grayish), spatial and temporal information can be inferred to reconstruct a \ac{PET} image. Modern scanners utilize \ac{TOF} information to estimate the annihilation point on the \ac{LOR}.}
\label{fig:PETSystem}
\end{figure}

\acresetall

\IEEEPARstart{A}{rtificial} intelligence (\acs{AI}) is finding its way more and more into medical imaging \cite{wagner_artificial_2019, carneiro_organizational_2018}, including the research field around \ac{PET} \cite{zhang_hierarchical_2023}. In contrast to \ac{CT} \cite{buzug_computed_2008} or \ac{MRI} \cite{vlaardingerbroek_magnetic_2003}, \ac{PET} is a functional imaging technique that does not reproduce anatomical structures but can visualize metabolic processes in the body. \ac{PET} uses the effect of electron-positron annihilation to obtain information about processes within the object of interest. A radioactive tracer is administered to the patient, accumulating in highly metabolic regions and emitting positrons \cite{gambhir_molecular_2002}. These positrons annihilate with the surrounding tissue, producing two $\gamma$-photons emitted back-to-back defining a \ac{LOR}. The $\gamma$-photons are subsequently registered in coincidence by a \ac{PET} scanner (see \cref{fig:PETSystem}) equipped with scintillation detectors, which convert the $\gamma$-photons into many optical photons in the visible light range that can be measured with a photosensor \cite{muehllehner_positron_2006}. Analog \acp{SiPM} utilize an \ac{ASIC} to digitize the signal pulses, whereas digital \acp{SiPM} perform digitization at a \ac{SPAD} level. Based on the detection information, especially the spatial and the temporal information, a \ac{PET} image can be reconstructed. While the application of neural networks in medical imaging usually focuses on the image reconstruction process \cite{gondara_medical_2016, jin_deep_2017, mahmud_applications_2018}, improvements along the complete imaging chain, from detecting physical signals \cite{van_den_ende_self-supervised_2023} to processing the resulting data, can improve the resulting image to facilitate medical diagnoses. In this work, we show the application of learning algorithms in the context of residual physics at the detector level to significantly improve the achievable \ac{CTR}. Especially in medical and physical applications, it is desired to get insight into the inner workings of models to ensure that the algorithms can capture meaningful relations. Therefore, we use eXplainable AI (XAI) \cite{samek_explainable_2017, samek_towards_2019, barredo_arrieta_explainable_2020, tjoa_survey_2021} methods to check whether trained models are able to understand simple physical constraints implied by the data generation.\newline
State-of-the-art clinical \ac{PET} scanners combine high spatial resolution with precise \ac{TOF} information (see \cref{fig:PETSystem}). Including the timing information in the image reconstruction process provokes an improvement in the \ac{SNR} of the image \cite{conti_new_2019} without increasing the radioactive dose and therefore improves also lesion detectability \cite{surti_update_2015}. Most \ac{PET} systems \cite{sluis_performance_2019, prenosil_performance_2022} utilize segmented scintillator topologies (see \cref{fig:SensorsAndDetectors}) due to the readout simplicity and very good timing performances. Contrary to this, (semi-)monolithic detector concepts spread the light over multiple channels. Recently, they have gained attention \cite{krishnamoorthy_performance_2018,gonzalez-montoro_new_2022} as they provide high spatial resolution \cite{muller_gradient_2018,he_preliminary_2021,decuyper_artificial_2021} but also offer intrinsic \ac{DOI} capabilities \cite{muller_novel_2019, loignon-houle_doi_2021}, thus, reducing parallax errors at reduced costs compared to segmented topologies. While, e.g., the $\gamma$-positioning strongly profits from the spread detection information, it creates disadvantages for the timing performance due to an enhancement of timewalk effects \cite{vinke_time_2010, du_time-walk_2017, xie_optical_2021} and jitter in signal-propagation times \cite{thompson_central_2004,Schug2015HitAnalysis}, which deteriorate \ac{CTR}. Therefore, monolithic detector concepts demand advanced readout algorithms and calibration routines to infer the needed information from the detected optical information.
Due to the light-spreading characteristic of \mbox{(semi-)monolithic} detectors, many approaches use machine or deep learning techniques, e.g., to infer the $\gamma$-interaction position within the scintillator volume. This strategy suggests itself since the detected optical photons represent abstract patterns that can easily be recognized by learning algorithms. However, applying machine learning for time skew calibration and estimation still remains experimentally a hard task since skew effects can vary in their magnitude and also in the incorporated nature on an event basis without the need for a spatial relation. Besides this, supervised learning demands labeled data, which is a priori not accessible without using simulation techniques, and unsupervised learning is often used in the context of clustering and association algorithms \cite{dike_unsupervised_2018} unsuitable for the proposed problem.
Recently, we proposed an analytical timing calibration technique \cite{naunheim_analysis_2022} suitable for traditional segmented and light-sharing-based scintillator topologies. This analytical calibration aims to reduce sequentially major skew effects by using a convex optimization of a matrix equation. When applying the technique, one observes that the skew effects are iteratively reduced. Within each iteration, the experimenter can address different characteristics of the time skews, e.g., by choosing a different separation into sub-volumes (called voxels) of the scintillation crystals. At a certain number of iterations, we see that the reported correction values $\hat{\vec{c}}$ oscillate around the baseline and that the \ac{CTR} does not improve further, indicating that the linear formulation of the problem, with $\boldsymbol{M}$ denoting the matrix and  $\vec{\overline{\Delta t}}$ the estimated mean time difference between the calibration objects,
\begin{equation}
\hat{\vec{c}} = \text{arg min}_{\vec{c}}   \|  \vec{\overline{\Delta t}} - \boldsymbol{M} \cdot \vec{c} \|_{_2},
\end{equation}
has limited capability of completely describing the physical situation. This challenge can theoretically be addressed by changing the mathematical formulation representing also the effects of higher order. However, this requires prior knowledge of the precise optical processes \cite{korzhik_physics_2020} taking place in the chosen scintillator topology, in order to change the mathematical formulation \cite{davi_brief_2021}. Furthermore, depending on the readout infrastructure and the detector concept, the problem might depend on numerous variables and parameters \cite{seifert_comprehensive_2012,cates_analytical_2015} which are hard to determine in advance. Hence, covering the effects of higher order can become arbitrarily complicated. In addition, detectors can vary in response such that an optimized representation might only fit the specific detector. A statistical approach using maximum likelihood was presented by Van Dam et al. \cite{van_dam_sub-200_2013}, focusing on differences between timestamps. We propose to use a machine learning approach instead and furthermore utilize a special way of experimental data generation to propose simple prior physical knowledge to the model by shifting a radiation source to different known positions \cite{naunheim_towards_2021}. We intend to apply this technique on top of the conventionally used analytical approach, forcing the algorithm to learn the effects of higher order, which we understand as residual physics \cite{zeng_tossingbot_2020, zhang_residual_2022}. By following this, we free ourselves from precisely modeling and catching all non-linear effects in the complete scintillation and detection process. In this work, we employed \ac{GTB} as learning algorithm since it is able to handle missing data \cite{mueller_semi-monolithic_2022} and allows usage in (near) real-time processing systems \cite{wassermann_high_2021} due to the simplicity of the model's architecture.\newline
The proposed approach is studied using experimental data acquired with a coincidence setup equipped with a semi-monolithic (\qtyproduct{3.9 x 31.9 x 19.0}{\milli \metre}) and a one-to-one coupled (\qtyproduct{3.9 x 3.9 x 19.0}{\milli \metre}) detector array concept. We trained multiple models on the acquired data and studied their performance based on the physics-related learning task, the agreement with theoretical expectations and bias effects, and the obtained \ac{CTR} values.

\section{Related Works}
\subsection{Approaches towards Residual Physics}
To the authors' knowledge, the first popular mention of 'residual physics' in the context of artificial intelligence was by Zeng et al. \cite{zeng_tossingbot_2020}. In their work, they investigated whether a robotic arm is able to pick up arbitrary objects and throw them into selected target boxes. While the problem of throwing can be described sufficiently well in theory by Newtonian physics, the real-world implementation for arbitrary objects is very challenging due to numerous additional variables that affect the throw.\newline
Similar works have been \cite{abbeel_using_2006, pastor_learning_2013, higuera_adapting_2017, silver_residual_2018, johannink_residual_2019} and are still being published \cite{kloss_combining_2022} in the context of 'hybrid controllers'. All of the studies having in common that they exploit the residuals between well-understood idealized physics and actual measurement.\newline
Alternative approaches aiming to combine physics domain knowledge and \ac{AI} are given by 'physics-informed learning' \cite{fu_novel_2022,oszkinat_uncertainty_2022,li_federated_2023,hua_physics-informed_2023}, where the utilized loss function is often modified to guide the model to physics-meaningful predictions.

\subsection{Timing Capabilities of PET Detectors}
In recent publications \cite{zhang_performance_2019, zhang_thick_2021, mueller_semi-monolithic_2022, freire_position_2022}, it has been shown that (semi-)monolithic detectors are able to provide good performances. Especially their timing capabilities have been studied under various experimental settings. Van Dam et al. \cite{van_dam_sub-200_2013} were able to reach sub-\qty{200}{\pico \second} \ac{CTR} for a monolithic crystal (\qtyproduct{24 x 24 x 20}{\milli \metre}) using a maximum likelihood approach and a measurement temperature of \qty{-20}{\degreeCelsius}, challenging to implement in a \ac{PET} system designed for the clinical domain. Sánchez et al. developed a new \ac{ASIC} (HRFlexToT \cite{sanchez_hrflextot_2022}) with redesigned energy measurement for linear \ac{ToT} behavior while reducing power consumption and improved timing response, and achieved \qty{324}{\pico \second} \ac{CTR} for a big monolithic crystal (\qtyproduct{25 x 25 x 20}{\milli \metre}). In a recent simulation study, Maebe et al. \cite{maebe_simulation_2022} reported \qty{141}{\pico \second}. In their simulation, they used a monolithic detector (\qtyproduct{50 x 50 x 16}{\milli \metre}) and a \ac{CNN}, while the network's input is given by the digitized waveforms truncated to a window of \qty{3}{\nano \second} using a step size of \qty{100}{\pico \second}.\newline
Zhang et al. \cite{zhang_thick_2021} reported a timing resolution of about \qty{718}{\pico \second} with thick semi-monoliths (\qtyproduct{1.37 x 51. 2 x 20}{\milli \metre}) digitized with the TOFPET2 ASIC at measurement temperatures of \qty{28}{\degreeCelsius}. Using energy-weighted averaging of timestamps reported by the TOFPET2 \ac{ASIC}, Cucarella et al. \cite{cucarella_timing_2021} achieved a \ac{CTR} of \qty{276}{\pico \second} for slabs with a volume of \qtyproduct{25.4 x 12 x 0.95}{\milli \metre}.\newline
In a proof-of-concept study performed by Berg et al. \cite{berg_using_2018} using two small lutetium fine silicate crystals (\qtyproduct{5 x 5 x 10}{\milli \metre}) coupled to a single photomultiplier tube, a timing resolution of about \qty{185}{\pico \second} was achieved using \acp{CNN}. Onishi et al. \cite{onishi_unbiased_2022} proposed a simple method for unbiased \ac{TOF} estimation by applying a combination of a \ac{CNN} and a \ac{LED} to an oscilloscope equipped with a pair of single scintillation \ac{LYSO} crystal of dimensions \qtyproduct{3 x 3 x 10}{\milli \metre} reaching \qty{159}{\pico \second}.

\section{Learning Algorithm \& Materials}
\subsection{Gradient Boosted Decision Trees}
\label{subsec:gtb}

While we utilized \ac{GTB} in this work, the presented calibration approach is also applicable to different learning architectures, e.g., deep neural nets. \ac{GTB} is a supervised learning algorithm based on an ensemble of binary decision trees, where each tree is trained on the residuals of the already established ensemble (additive training). In this work, we use the \ac{GTB} implementation of XGBoost \cite{chen_xgboost_2016}, with the model $\phi$,
\begin{equation}
\phi = \sum \limits_{k=1}^K f_k, 
\end{equation}
being given as the superposition of the $K$ trees (weak learners) $f_k$. Each tree $f_k$ is an element in the CART \cite{breiman_classification_2017} space $\Omega$,
\begin{equation}
f_k \in \Omega.
\end{equation}
In its design, \ac{GTB} is a relatively simple architecture compared to widely used deep neural networks \cite{tang_extreme_2016, wu_comprehensive_2021, li_survey_2022}. However, it has proven high predictive power in many applications \cite{xia_boosted_2017, torlay_machine_2017, aaboud_observation_2018,zhang_gbdt-mo_2021, shwartz-ziv_tabular_2022}, and due to its simplicity, \ac{GTB} allows usage in high throughput software \cite{wassermann_high_2021} suitable for complete \ac{PET} systems or even the application directly on the detector level \cite{shepovalov_fpga_2020, alcolea_fpga_2021}. Regarding the scope of this work, two hyperparameters of \ac{GTB} models are of particular importance, namely the maximal depth $d$, denoting the maximal number of decisions within an ensemble, and the learning rate $lr$, measuring the residual influence on the learning of the following tree. The learning rate must be optimized in most cases to find a compromise between training duration and accuracy. A third prominent hyperparameter is the number of trees $n$ of an ensemble. We excluded $n$ from the hyperparameter search in this work since we used an early stopping criterion.

\subsection{Shapley Additive Explanations}
The SHAP (SHapley Additive exPlanations) framework \cite{Lundberg2017, Lundberg2018} is used as an explainable AI technique to analyze feature importance in order to search for correlations between physical effects and patterns the model has learned. In particular, in this work, we utilized the TreeExplainer implementation \cite{lundberg_local_2020} because of the chosen learning architecture. The framework uses mathematical game theory. Each input sample and corresponding prediction is connected by assuming a coalition game. The players in the game are represented by the feature values of the input sample, where each feature influences the model's prediction. These influences are called contributions and are expressed in the same physical unit as the predictions. Contribution values are mathematically either positive or negative, while the model's output is equal to the sum over the contributions. The magnitude of a given contribution indicates the level of its importance.\newline
SHAP uses Shapley values \cite{Shapley+2016+307+318}, which are a measure to quantify the contribution of a feature regarding the specific model's output. In a mathematical sense, SHAP combines three concepts essential for providing a consistent picture concerning feature importance. Firstly, the SHAP values must satisfy local accuracy, meaning that for a given input sample, the sum of the estimated feature contributions must be equal to the corresponding model's prediction that should be explained. If a feature is missing, it cannot attribute to importance, which is covered in SHAP using the concept of missingness. Lastly, consistency is required, ensuring that when changing a model such that a particular feature has a larger impact on the model, the corresponding attribution cannot decrease.\newline
Practically, for each feature value $f_k$ of a given input sample $X = \{ f_k \}$, an associated SHAP value $SV(f_k)$ can be computed, reporting a local explanation that connects the feature value with its contribution to the model's output $y$. By combining many local explanations, one can conclude a global understanding of the model.\newline

\subsection{PET Detectors}

\begin{figure}
\centering
\includegraphics[width=\scBroad cm]{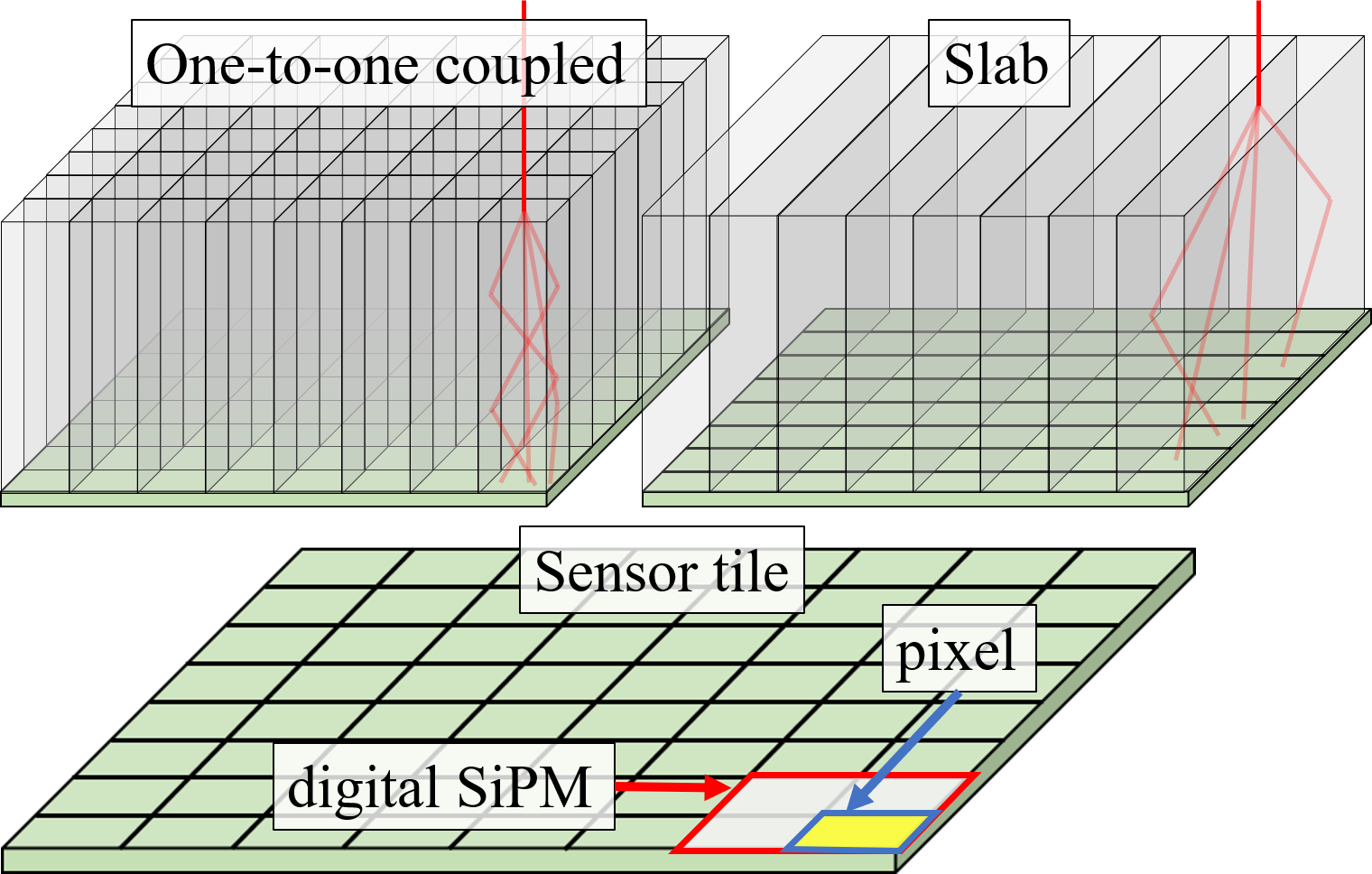}
\caption{Used scintillator topologies and photosensors. The incoming $\gamma$-photon, as well as a part of the optical photons are illustrated as red lines. The sensor tile consists of \numproduct{4x4} digital \acp{SiPM} (DPC3200-22, Philips Digital Photon Counting), each one holding four pixels and a twin \acl{TDC}. A triggered \ac{SiPM} reports four pixel count values and a timestamp.}
\label{fig:SensorsAndDetectors}
\end{figure}

The study is conducted using two different detector types, where one detector is based on a one-to-one coupled scintillator design, and the other detector is based on a semi-monolithic scintillator design (see \cref{fig:SensorsAndDetectors}).\newline
Each scintillator concept is glue-coupled (Meltmount, Cargille Laboratories) to a sensor tile holding \numproduct{4x4} digital \acp{SiPM} (DPC3200-22, Philips Digital Photon Counting, Aachen \cite{Frach2010}). Each \ac{SiPM} is formed by \numproduct{2 x 2} readout channels (also called pixels) and a twin \acl{TDC}, where one readout channel consists of \num{3200} \acp{SPAD}. Each \ac{SiPM} of a sensor tile works independently and follows a configured acquisition sequence if a pre-defined internal two-level trigger scheme is fulfilled. After the reception of a trigger, it is checked during the validation phase if the geometrical distribution of discharged \acp{SPAD} met the configured requirement. If both trigger thresholds are fulfilled, the acquisition is continued. Each triggered \ac{SiPM} provides information that encloses a timestamp and four pixel photon count values, called a hit.\newline
Both scintillators use LYSO as scintillation material (Crystal Photonics, Sanford). Concerning the scintillator architecture, an array of \numproduct{8x8} LYSO segments of \qty{4.0}{\milli \metre} pitch and \qty{19.0}{\milli \metre} height is utilized in the one-to-one coupled design. Each segment is wrapped with \ac{ESR} foil and covers the pitch of one pixel.\newline
The semi-monolithic detector concept comprises eight monolithic LYSO slabs, each having a volume of \qtyproduct{3.9 x 31.9 x 19.0}{\milli \metre}. Each slab aligns with one row of pixels. \ac{ESR} foil is located between every second slab and on the laterals walls to reduce light sharing between trigger and readout regions. The slab detector is able to provide intrinsic \ac{DOI} information due to its monolithic characteristics. Not all \acp{SiPM} that are partly covered by a slab might be triggered and send hit data corresponding to a $\gamma$-interaction due to the independent operation of the \acp{SiPM}.

\subsection{Coincidence Setup}
\begin{figure}
\centering
\includegraphics[width=\scBroad cm]{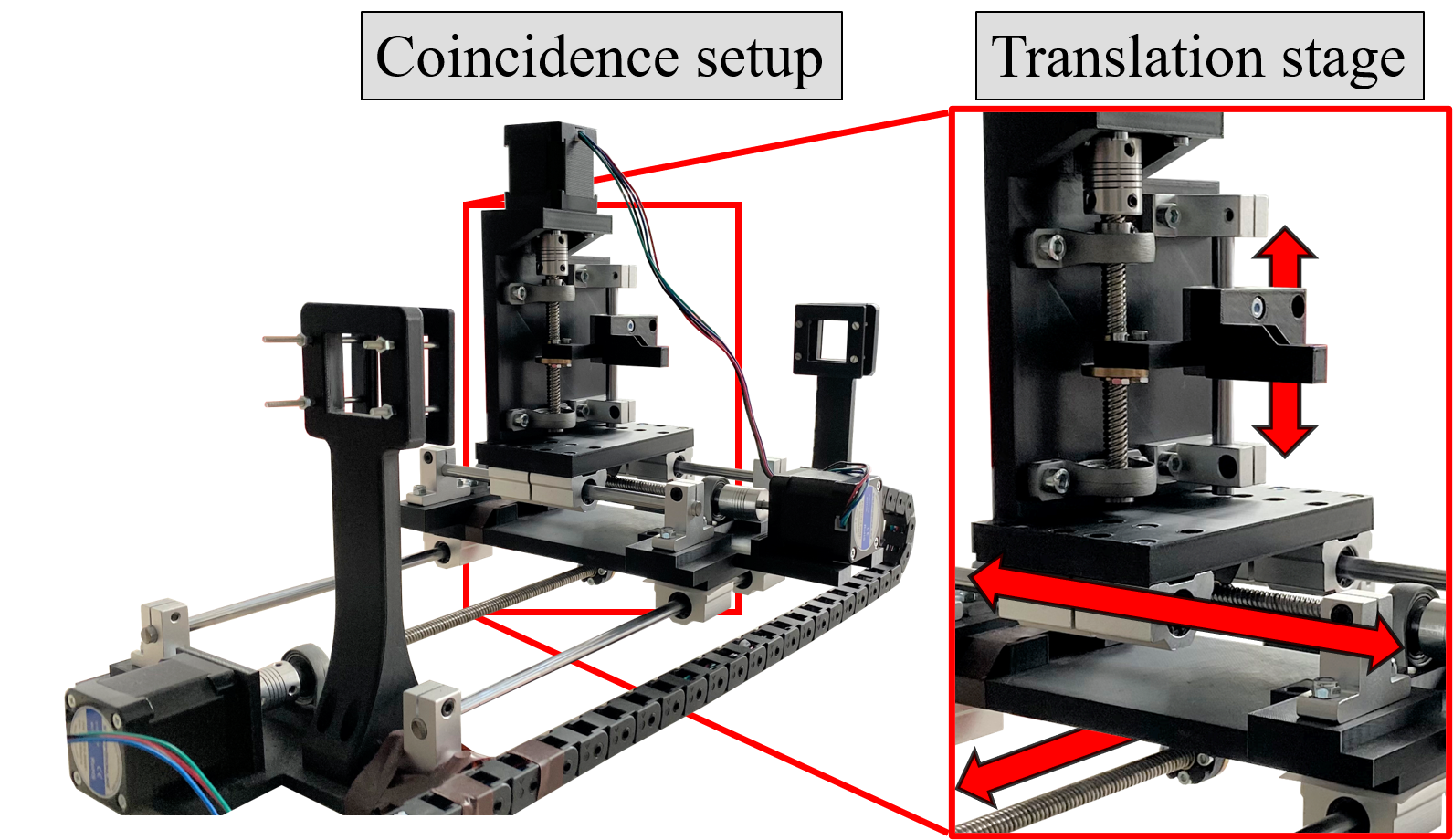}
\caption{Used coincidence setup for the acquisition of labeled data. The source mounting is connected to the translation stage system allowing motion along all three axes (indicated as red arrows).}
\label{fig:TranslationStage_full}
\end{figure}

The experimental setup comprises a source mounting, in addition to the detectors, and is located in a tempered dark box. The source mounting is connected to a programmable translation stage system, allowing motion in all three spatial axes (see \cref{fig:TranslationStage_full}). The distance between the detector surfaces is given to be \qty{435}{\milli \metre}. The precision of the translation stage considering the complete measurement range is given to be \qty{10}{\micro \metre} which translates regarding a coincidence measurement to an uncertainty in the time domain of about \qty{0.067}{\pico \second}. The source mounting is equipped with a $^{22}$Na source with an activity of approximately \qty{12}{\mega \becquerel} and a diameter of \qty{0.5}{\milli \metre}. Coincidences are acquired by utilizing flood irradiation and moving the source to various positions between the detectors.

\section{Experiments}
\subsection{Data Acquisition}

The proposed calibration technique uses supervised machine learning and therefore demands labeled data. The labels are generated by moving the radiation source to specific positions between the facing detectors and measuring coincidences. Thanks to the known source position, one can calculate the expected time difference $\mathbb{E}(\{\Delta t\})$ because of the different path lengths the $\gamma$-photons have to travel until reaching the detector. The source was moved to \num{47} different $z$-positions (see \cref{fig:scheme_data_generation}) with a step size of \SI{5}{\milli \metre} ranging from \qtyrange{-130}{100}{\milli \metre}, while at each $z$-position, a grid of \numproduct{5 x 5} positions in the $xy$-plane with a step size of \SI{6}{\milli \metre} was utilized to acquire coincidences. Both, $x$ and $y$ positions ranged from \qtyrange{-12}{12}{\milli \metre}. At each grid point a measurement time of \qty{600}{\second} was set, resulting in a total measurement time of about \qty{8}{\day}. The acquired measurement data consisting of \num[round-mode=places, round-precision=2, exponent-mode=scientific]{682351456} coincidences ($\approx \num[round-mode=places, round-precision=2, exponent-mode=scientific]{580425.5319148937}$ coincidences per position) is finally used to form three datasets for training, validation, and testing during the model-building process comprising \num[round-mode=places, round-precision=2, exponent-mode=scientific]{329000000}, \num[round-mode=places, round-precision=2, exponent-mode=scientific]{155951456}, and \num[round-mode=places, round-precision=2, exponent-mode=scientific]{197400000} input samples, respectively. We decided to evaluate the final \ac{CTR} performance (see \cref{subsec:CTR_per_res}), using data from a measurement conducted at a different day using the same conditions and detectors to prove the predictive power and generalized applicability of the trained models. This dataset comprises \num[round-mode=places, round-precision=2, exponent-mode=scientific]{4200631} coincidences acquired with the radiation source located near the iso-center of the setup, as it is usually done for \ac{CTR} evaluations. To allow a clean separation in the naming, the dataset used for the \ac{CTR} performance evaluation is called performance dataset, while the three other datasets remain in the usual naming (training, validation, testing).\newline
During both acquisitions, the sensor tile reported a constant temperature of \SI{2.1}{\degreeCelsius} for the one-to-one coupled detector and \SI{0.0}{\degreeCelsius} for the slab detector. Both sensor tiles were operated in first-photon trigger \cite{Tabacchini2014}. The excess voltage was adjusted to \SI{2.8}{\volt}, while the validation pattern was set to scheme \num{16} (0x55:AND) demanding on average \num{54 \pm 19} optical photons \cite{PhilipsDigitalPhotonCounting2016}.

\subsection{Data Pre-Processing \& Preparation}
\subsubsection{Coincidence Clustering}
Data associated with one $\gamma$-interaction has to be clustered due to the independent readout of the DPCs. A cluster window of \qty{40}{\nano \second} is reasoned by the timestamp difference distribution of the hits’ uncorrected timestamps to combine all hits into a cluster correlated to the same $\gamma$-interaction. Measured raw data were corrected for saturation effects, and the \aclp{TDC} of each DPC were linearly calibrated against each other, assuming a uniform distribution of triggers regarding a clock cycle \cite{Schug2015HitAnalysis}. Clusters with less than \num{400} or more than \num{4000} detected optical photons were rejected for noise reduction since the non-calibrated photopeak of the \qty{511}{\kilo \eV} $\gamma$-photons was located at \num{2300} and \num{2800} for the slab and one-to-one coupled detector, respectively. Coincidences were grouped on cluster level using a sliding coincidence window of \SI{10}{\nano \second} considering the first timestamp of two clusters.

\subsubsection{Position \& Energy Estimation}
\label{subsec:pos_energy_estimation}

A subset of the features used during the proposed time skew calibration is given by the $\gamma$-interaction position inside the scintillator volume and by the deposited and calibrated energy in units of \unit{\kilo \eV}. To acquire the positioning and energy information of each event, dedicated calibrations already established in previous works \cite{muller_gradient_2018, muller_novel_2019} were performed.\newline
While the $\gamma$-positioning in the one-to-one coupled detector is given by the pixel's position showing the highest photon count, the semi-monolithic slab detector requires a calibration procedure to estimate the $3$D interaction location. For this purpose, \ac{GTB} \cite{chen_xgboost_2016, muller_gradient_2018, muller_novel_2019} models are trained based on data acquired with an external reference using a fan-beam setup \cite{Hetzel2020}, which irradiates the scintillator at known positions. While the positioning resolution of the one-to-one coupled detector is given to be \SI{2}{\milli \metre}, the slab detector's resolution is in the planar direction \qty{2.5}{\milli \metre} and in the \ac{DOI} direction \qty{3.3}{\milli \metre}. The positioning resolution is determined by the \ac{FWHM} of the positioning error distribution \cite{muller_novel_2019}.\newline
The energy value associated with a $\gamma$-interaction is estimated using a $3$D-dependent energy calibration utilizing an averaged light pattern. The crystal volume is divided into $n_x \times n_y \times n_{doi}$ voxels, where for each voxel, the mean number of detected optical photons is estimated, based on $\gamma$-events, whose interaction positions were located inside the voxel volume. The slab detector is divided into \numproduct{8 x 8 x 4} voxels, while the one-to-one coupled detector is divided into \numproduct{8 x 8 x 1} voxels. The energy resolution of the one-to-one coupled detector was evaluated at \qty{10.4}{\percent}, while the energy resolution of the slab was estimated to be \qty{11.3}{\percent}.

\subsubsection{Analytical Timing Calibration}
\label{subsec:ana_timing}
The first part of the calibration is given by performing an analytical calibration, which has been studied in previous publications \cite{Schug2015HitAnalysis, mann_computing_2009, reynolds_convex_2011} and relies on well-known mathematical principles like convex optimization. In this work, our calibration formalism \cite{naunheim_analysis_2022} was used. However, the principle of exploiting residual physics remains also functional for every other analytical calibration.\newline
During the calibration process, multiple sub-calibrations are conducted, where in each sub-calibration different hyperparameters are applied such that one tries to address many aspects of time skew effects. The same convex optimization process is used within each sub-calibration in order to find suitable corrections $\hat{\vec{c}}$,
\begin{equation}
\hat{\vec{c}} = \underset{\vec{c}}{\text{arg min}}   ||\vec{\overline{\Delta t}} - \boldsymbol{M} \cdot \vec{c}||^2,
\end{equation}
with $\vec{c}$, and $\vec{\overline{\Delta t}}$, denoting the calibration channel vector and the mean time difference vector, respectively, and $\boldsymbol{M}$ encoding different channel combinations in the form of a matrix. After some number $N$ of performed sub-calibrations, a convergence of the detector \ac{CTR} value as well as the estimated corrections $\{ \hat{c}_k \} \in \hat{\vec{c}}$ is visible,
\begin{align}
\text{CTR}_i  \rightarrow \text{const.} & \text{ for } i \rightarrow N \\
\{ \hat{c}_k \}_i  \rightarrow \qty{0}{\pico \second} & \text{ for } i \rightarrow N,
\end{align}
with $i$ denoting the number of applied sub-calibration. For this work, we used three sub-calibration iterations (based on \ac{TDC} regions, readout channels, and voxels as described in \cite{naunheim_analysis_2022}) mainly addressing fixed skews due to differences in the signal propagation and time jitter introduced by the scintillator itself. At this point, it becomes inconvenient to add more and more sub-calibrations since the benefit decreases strongly.

\subsection{Residual Timing Calibration}

\begin{figure}
\centering
\includegraphics[width=\scBroad cm]{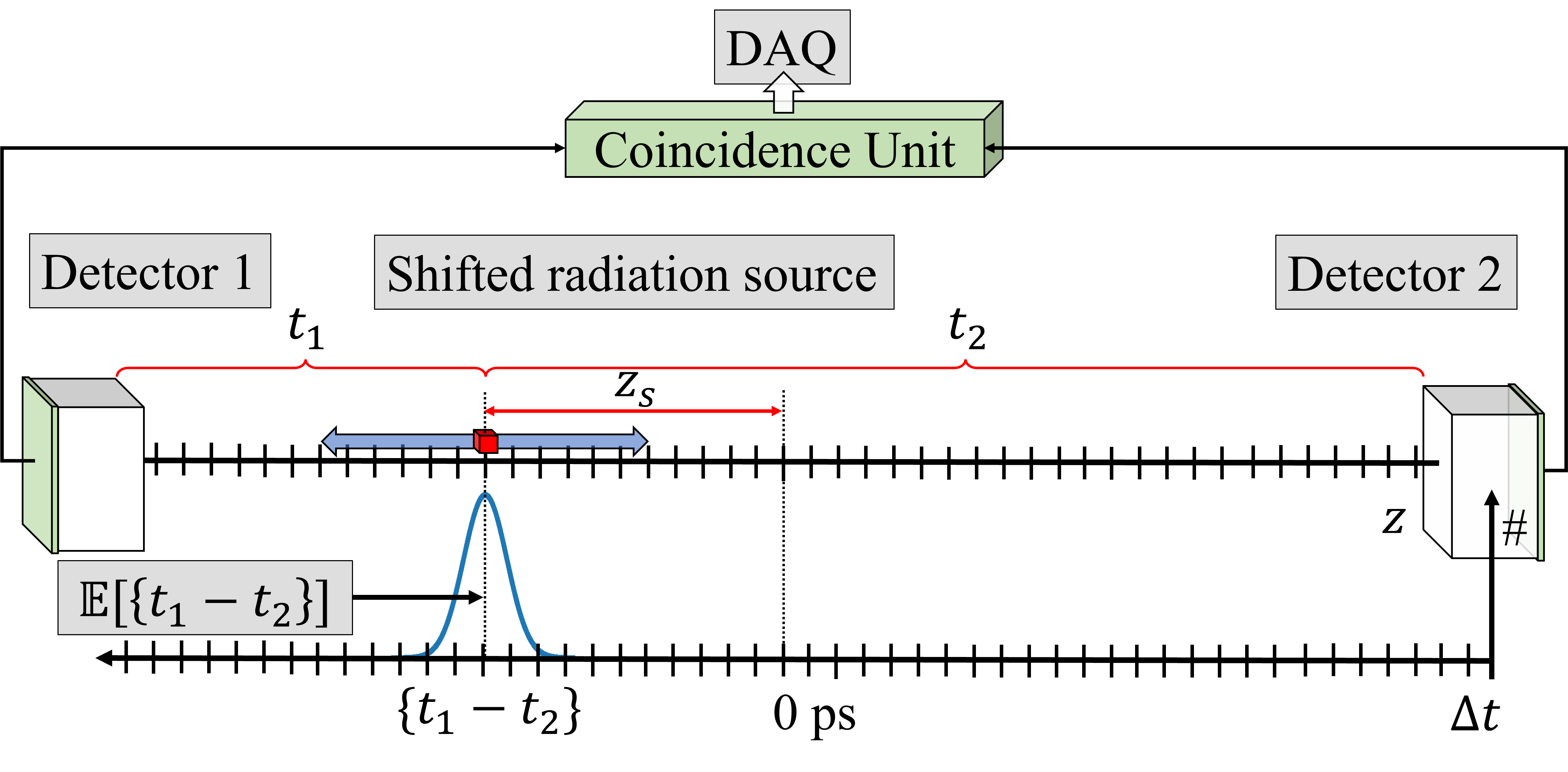}
\caption{Scheme of the labeling process to acquire data that can be used for supervised learning. The radiation source (red cube) is shifted to different positions $z_{\text{s}}$ along the centered coordinate system $z$. Varying source positions lead to different travel times $t_1$ and $t_2$ of the $\gamma$-photons. The expected time difference $\mathbb{E} [ \, \{ t_1 - t_2 \} ] \,$ is used as label for the learning process.}
\label{fig:scheme_data_generation}
\end{figure}

\begin{figure*}[t]
\centering
\includegraphics[width=16 cm]{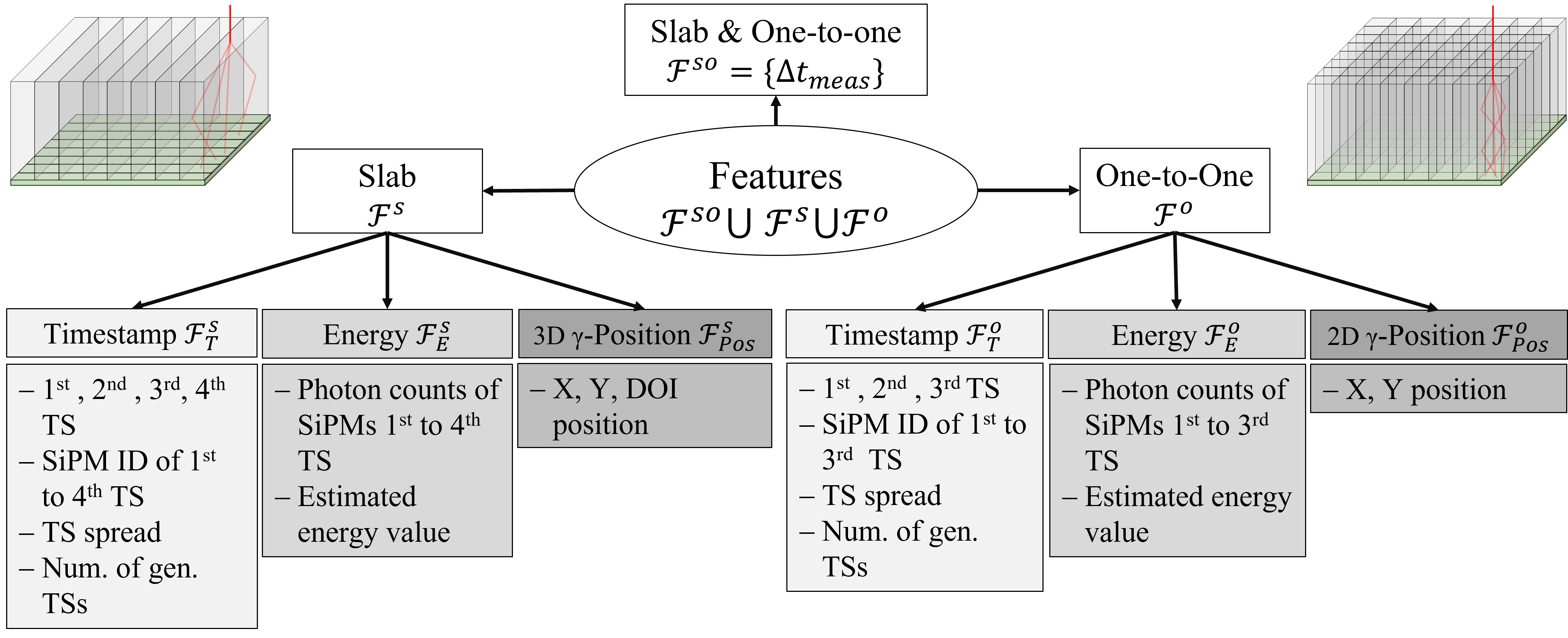}
\caption{Overview of the used features $\mathcal{F}$ to train the \ac{GTB} models. There are three different feature sets, given by purely slab detector-related features $\mathcal{F}^{\text{s}}$, purely one-to-one detector-related features $\mathcal{F}^{\text{o}}$, and features associated with both detector concepts $\mathcal{F}^{\text{so}}$. $\mathcal{F}^{\text{so}}$ consists only of the difference $\Delta t_{\text{meas}}$ between the first timestamps from slab and one-to-one coupled detector, respectively. The sets $\mathcal{F}^{\text{s}}$ and $\mathcal{F}^{\text{o}}$ are symmetrical in their content and can again be grouped into the subsets timestamp information $\mathcal{F}_{\text{T}}^{\text{s/o}}$, energy information $\mathcal{F}_{\text{E}}^{\text{s/o}}$ and position information $\mathcal{F}_{\text{Pos}}^{\text{s/o}}$. Information about the processed timestamps (denoted as TS), the SiPM IDs of those timestamps, the timestamp spread (difference between first and last timestamp of a cluster), and the number of generated timestamps is given. The latter equals also the number of hits within the cluster. Besides this, the photon counts of the corresponding SiPMs, the calibrated energy value and the spatial interaction position are used.}
\label{fig:overview_features}
\end{figure*}

We propose to use a data-driven approach on top of the conventionally used technique to explore new corrections that have not been covered by the analytical formulation and improve the \ac{CTR}. A suitable way of doing this is by using artificial intelligence to search for patterns in the acquired coincidence data. We decided to employ the supervised algorithm \ac{GTB} (see \cref{subsec:gtb}), which was also used during the $\gamma$-positioning (see \cref{subsec:pos_energy_estimation}). Using a supervised approach demands labeled data (input samples and corresponding target values known as labels) to train a model. However, for the proposed problem of non-static time skew effects labeling is difficult, since it is a priori not clear how many and how strong the worsening effects are pronounced in each measured coincidence. Using an analytical estimator to generate the ground truth would limit the capabilities of the trained model to the chosen estimator. In order to solve the problem of labeling, we propose to shift the radiation source to different positions and measure coincidences between the facing detectors. The $\gamma$-photons travel varying path lengths to the detectors resulting in different expected time differences per source position.\newline
The different path lengths of the $\gamma$-photons (see \cref{fig:scheme_data_generation}), lead to different travel times $t_1$ and $t_2$. One can conclude the expected time difference $\mathbb{E} [ \, \{ t_1 - t_2 \} ] \,$, which is subsequently used as label $y$,
\begin{equation}
\label{eq:linear_label}
y = \mathbb{E} [ \, \{ t_1 - t_2 \} ] \ \approx \frac{-2 z_{\text{s}}}{c_{\text{air}}},
\end{equation}
with $c_{\text{air}}$ denoting the speed of light in air and $z_{\text{s}}$ denoting the source offset under the assumption that the coordinate system $z$ is located at the iso-center of the setup (see \cref{fig:scheme_data_generation}). For Gaussian distributions, the expectation value $\mathbb{E}$ is identical to the mean value of the distributions. Data acquired with the aforementioned scheme is further processed and finally used to train \ac{GTB} models. The input features $\mathcal{F}$ can be grouped into three categories: purely slab detector-related features $\mathcal{F}^{\text{s}}$, purely one-to-one detector-related features $\mathcal{F}^{\text{o}}$, and features associated with both detector concepts $\mathcal{F}^{\text{so}}$. While $\mathcal{F}^{\text{so}}$ consists only of the difference $\Delta t_{\text{meas}}$ between the first timestamps from slab and one-to-one coupled detector, respectively, $\mathcal{F}^{\text{s}}$, and $\mathcal{F}^{\text{o}}$ can be separated into the subsets timestamp information $\mathcal{F}_{\text{T}}^{\text{s/o}}$, energy information $\mathcal{F}_{\text{E}}^{\text{s/o}}$ and position information $\mathcal{F}_{\text{Pos}}^{\text{s/o}}$(see \cref{fig:overview_features}).

Since the detector-specific feature sets $\mathcal{F}^s$ and $\mathcal{F}^o$ are symmetrical in their content, we will explain the specific features in a generalized way in the following. The subset timestamp information $\mathcal{F}_{\text{T}}^{\text{s/o}}$ contains the four (three) first timestamp values reported by the slab (one-to-one coupled) detector. A trade-off between available information and needed memory reasons for the choice of the different number of used timestamps. For both detectors, the cumulative distribution of the number of generated timestamps per cluster was analyzed and determined to the value matching roughly \qty{80}{\percent} of all clusters. Let $\mathcal{T}_j$ be the set of timestamps provided within a cluster $j$ by the photodetector,
\begin{equation}
\mathcal{T}_j = \{ t_{j,0}, t_{j,1}, \ldots, t_{j,i}, \ldots \},
\end{equation}
with $t_{j,i}$ denoting the $i$-th timestamp of cluster $j$. Since the photosensor consecutively reports the timestamp values throughout the measurement, they need to be processed after the coincidence search to be suitable for feeding into a machine learning algorithm. Therefore, the very earliest timestamp $t_{j,0}$ of a cluster $j$ is subtracted from the following timestamps $t_{j,i}$ of this cluster,
\begin{equation}
\tilde{t}_{j,i} = t_{j,i} - t_{j,0},
\end{equation}
with $\tilde{t}_{j,i}$ denoting the processed timestamp $i$ of cluster $j$ employed as input.\newline
Furthermore, the origin of the respective timestamps is used and represented by their \ac{SiPM} ID number. Besides this, information about the cluster's timestamp spread (the difference between the first and last timestamp) and the number of timestamps (equals the number of hits) in the cluster is utilized. The subset energy information $\mathcal{F}_{\text{E}}^{\text{s/o}}$ contains information about the deposited energy as estimated energy value in \unit{\kilo \eV}, and as raw photon counts that have been detected on the corresponding \acp{SiPM}. The $\gamma$-positioning set $\mathcal{F}_{\text{Pos}}^{\text{s/o}}$ holds information about the interaction position of the $\gamma$-photon within the scintillator volume. While this is given as a $3$D position for the semi-monolithic case, the one-to-one coupled design provides only planar ($2$D) information.\newline
In order to find suitable hyperparameters regarding the learning task, a grid search was conducted considering the maximal tree depth $d$ and the learning rate $lr$, with
\begin{align}
d &\in \{ 12, 15, 18, 20 \}\text{, and} \\
lr &\in \{ 0.1, 0.3, 0.5 \}.
\end{align}
During the model-building process, the maximal number of estimators $n$ within an ensemble was set to $n=500$, where the final number of used estimators was defined by the built-in early stopping criterion considering ten early stopping rounds to suppress possible overfitting. The learning task is performed using XGBoost's default squared error loss function \cite{chen_xgboost_2016}.\newline

\subsection{MAE Evaluation \& Linearity of Predictions}

The \ac{MAE} is used to evaluate the performance of a trained \ac{GTB} model based on the testing data,
\begin{equation}
\text{MAE}(z_{\text{s}}) = \frac{\sum_{i=0}^N | y_i (z_{\text{s}}) - \hat{y}_i (z_{\text{s}}) |}{N},
\end{equation}
with $y_i (z_{\text{s}})$ denoting the label of sample $i$ belonging to the source position $z_{\text{s}}$, and $\hat{y}_i (z_{\text{s}})$ denoting the corresponding model prediction. We utilize information about the test data prediction distributions to verify their Gaussian shape using a goodness-of-fit approach and to validate that the linearity condition given by \cref{eq:linear_label} is fulfilled. This validation ensures that the trained models obey the physical principle and do not compress the time differences since it would artificially improve the \ac{CTR}. Therefore, a linear regression is performed for each trained \ac{GTB} model and each grid point $(x_{\text{s}}, y_{\text{s}})$ in a range from \qtyrange[fixed-exponent=0]{-75}{45}{\milli \metre}, considering the fitted mean value of the prediction distributions $\mu_{\text{s}}$ and the associated source position $z_{\text{s}}$. We assumed a linear dependence following 
\begin{equation}
\label{eq:linearity_cond1}
\mu_{\text{s}}(z_{\text{s}}| \varepsilon, b)  = \frac{-2}{c_{\text{air}}} \cdot \varepsilon \cdot z_{\text{s}} + b,
\end{equation}
while in theory
\begin{equation}
\label{eq:linearity_cond2}
\varepsilon \overset{!}{=} 1.
\end{equation}
All fitting procedures are performed using SciPy's ODR package \cite{noauthor_orthogonal_nodate}. The uncertainty $\sigma_{\mu_{\text{s}}}$ on $\mu_{\text{s}}$ was based on the uncertainty on the mean reported by the fit procedure. Furthermore, an uncertainty on the translation stage position was given to be the same for all source positions \mbox{$\sigma_{z_{\text{s}}} = \qty{0.1}{\milli \metre}$}. Finally, the global linearity performance is given by the averaged $\varepsilon$-value for each model.

\subsection{CTR Performance}

To evaluate the timing performance, the \ac{FWHM} of the predicted time difference distribution is estimated by fitting a Gaussian function. The error on the estimated timing resolution is given by the uncertainty on the fitted \mbox{$\sigma$-parameter} of the Gaussian. The input data is given by the performance dataset. The \ac{CTR} is estimated for unfiltered data, for coincidences within a large energy window from \qtyrange{300}{700}{\keV}, and for coincidences within a smaller energy window from \qtyrange{450}{550}{\keV}.

\subsection{SHAP Analysis}

Due to computational costs, the SHAP analysis was performed for the model showing the best \ac{MAE} and \ac{CTR} performance using a subset of \num{23500} samples of the performance data. The analysis was done without applying any filters.

\begin{figure}
\centering
\includegraphics[width=\scBroad cm]{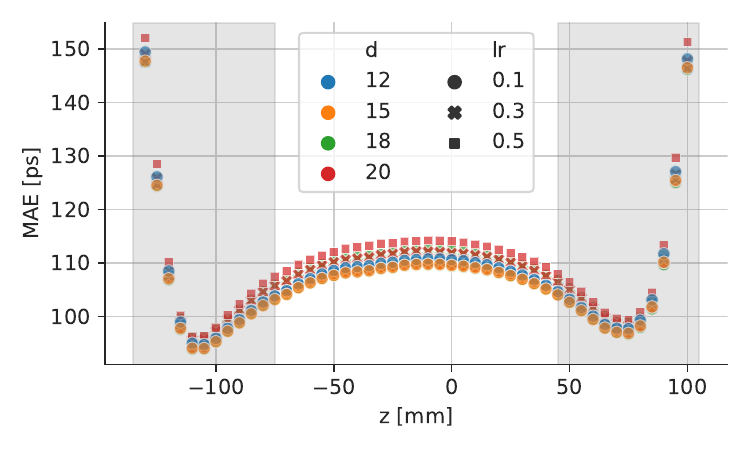}
\caption{Progression of the \ac{MAE} for each source position $z_{\text{s}}$ contained in the test dataset. No energy filter or restrictions on the measured light distribution were applied. Models utilizing a small learning rate show the best performances. Predictions located in the grayish areas \mbox{($z \notin [-75, 45] \unit{\milli \metre}$)} are excluded from the linearity analysis since the \ac{MAE} progression indicates the starting of the transition into the artifact-dominated region for these points.}
\label{fig:mae_performance}
\end{figure}

\section{Results}
\subsection{MAE Evaluation \& Linearity of Predictions}

\begin{figure*}[b]
\centering
\includegraphics[width=16 cm]{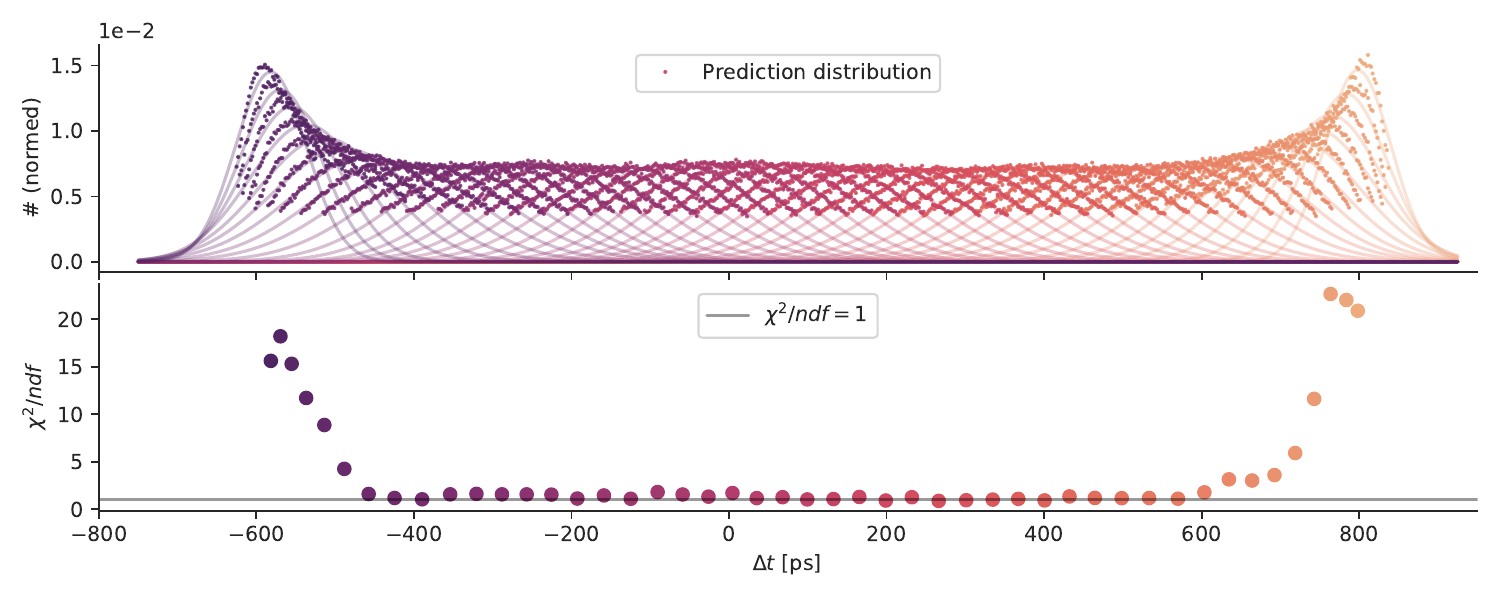}
\caption{Depiction of the distribution of predictions of the model using the hyperparameters $(d=18,lr=0.1)$ using all \num{47} source positions along the $z$-axis. The different source positions $z_{\text{s}}$ are encoded in color. The upper plot shows the different histograms and Gauss fits. The lower plot shows the goodness-of-fit per number of degrees of freedom ($\chi^2/ndf$) value for each position. For a large central region, the predictions are in very good agreement with a Gaussian function. When moving toward the edges of the data, the distribution becomes skewed and deviates from the Gaussian shape. No energy filter or restrictions on the measured light distribution were applied.}
\label{fig:overview_chiq}
\end{figure*}

The \ac{MAE} performance (see \cref{fig:mae_performance}) is similar for all chosen hyperparameter configurations. The distribution shows a symmetrical behavior around the median value of $\tilde{z}= \qty{-15}{\milli \metre}$, with a slight skewness that can be observed going from negative offset positions toward positive ones. While the prediction quality strongly decreases at the borders of the presented data, the models' predictions work very well in the central region. In general, one observes that models with a lower learning rate perform slightly better than those with a learning rate equal to or higher than $0.3$. Furthermore, \cref{tab:mae_overview} reveals that the \ac{MAE} is reduced by restricting the allowed energy of the test data. The model with hyperparameter configuration \mbox{$(d=18,lr=0.1)$} achieved the best \ac{MAE} performance. For the linearity analysis, we excluded the predictions located outside an interval of \qty{\pm 60}{\milli \metre} around the median (grayish areas in \cref{fig:mae_performance}) to be able to give an unbiased evaluation of the performance in the large central region of the data.

\begin{table}
\centering
\caption{Overall \ac{MAE} performance for different hyperparameter configurations and energy windows of the test data.}
\label{tab:mae_overview}
\begin{tabular}{@{}llll@{}}
\toprule
 & \multicolumn{3}{c}{MAE {[}ps{]}} \\ 
\multirow{-2}{*}{Model $(d,lr)$} & all & $[300,700]$ keV & $[450,550]$ keV \\ \cmidrule(r){1-4}
(12, 0.1) & 107.86 & 85.17 & 80.18 \\
(12, 0.3) & 107.50 & 84.90 & 79.75 \\
(12, 0.5) & 108.32 & 85.61 & 80.41 \\
(15, 0.1) & 106.92 & 84.40 & 79.32 \\
(15, 0.3) & 107.79 & 85.07 & 79.91 \\
(15, 0.5) & 108.84 & 85.99 & 80.80 \\
\rowcolor[HTML]{C0C0C0} 
(18, 0.1) & 106.87 & 84.23 & 79.09 \\
(18, 0.3) & 108.40 & 85.42 & 80.23 \\
(18, 0.5) & 109.47 & 86.23 & 80.93 \\
(20, 0.1) & 107.30 & 84.45 & 79.24 \\
(20, 0.3) & 109.02 & 85.66 & 80.30 \\
(20, 0.5) & 110.76 & 87.07 & 81.76 \\ \bottomrule
\end{tabular}
\end{table}

\begin{figure}
\centering
\includegraphics[width=\scBroad cm]{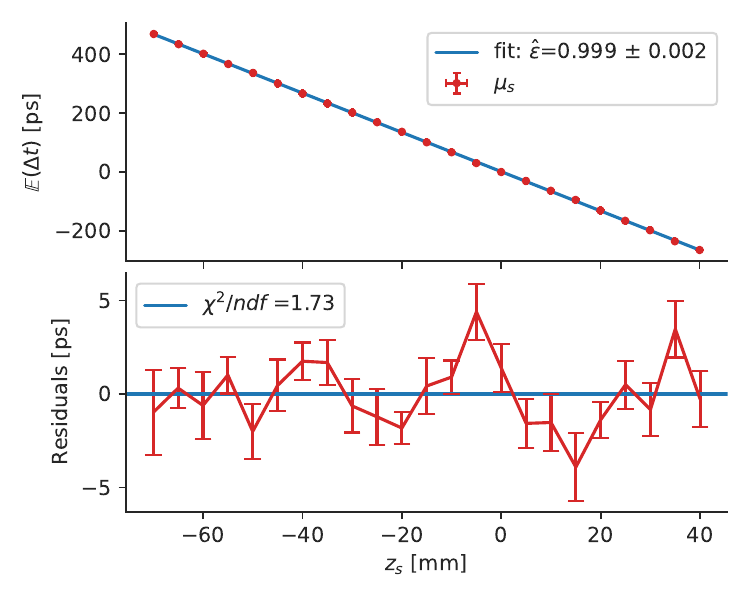}
\caption{Linear regression and residual plot of the linearity analysis of model \mbox{$(18, 0.1)$} for \mbox{$(x_{\text{s}},y_{\text{s}}) = (12,0)\unit{\milli \metre}$}. The Gaussian fitting procedure gives the uncertainty on $\mu_{\text{s}}$, while the uncertainty on the $z_{\text{s}}$ position is assumed to be \qty{0.1}{\milli \metre}.}
\label{fig:example_linReg}
\end{figure}

\begin{figure}
\centering
\includegraphics[width=\scBroad cm]{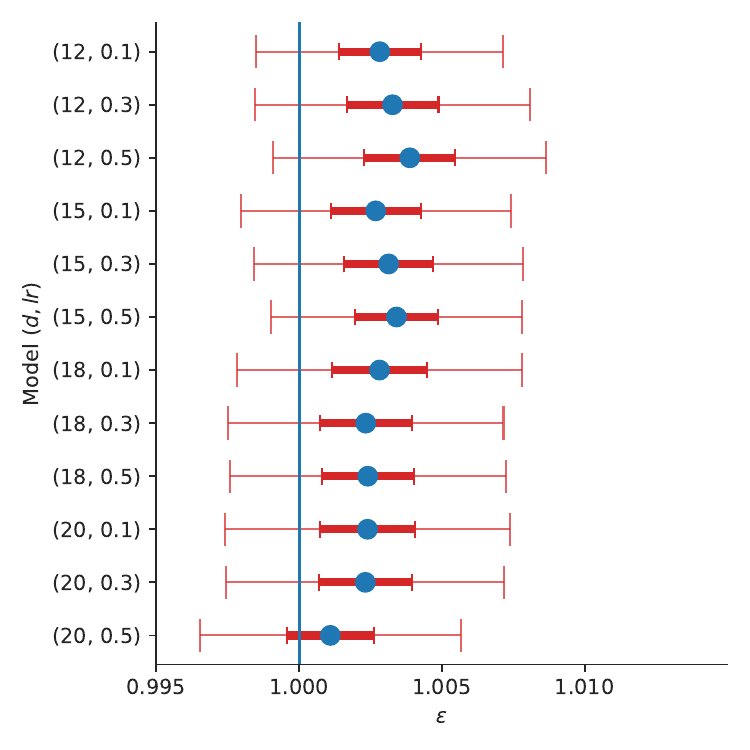}
\caption{Global linearity analysis for each trained model. The $\varepsilon$-values are based on the test dataset applying an energy window of \qtyrange{300}{700}{\kilo \eV}. The thick errorbars represent $1\sigma$-deviation, while the thin errorbars illustrate $3\sigma$-deviation.}
\label{fig:epsilon_perf1}
\end{figure}

\Cref{fig:overview_chiq} shows exemplarily the distribution of the predictions considering the complete data range for the model $(18, 0.1)$ in combination with the goodness-of-fit per number of degrees of freedom ($\chi^2/ndf$) for a Gaussian function. Both distributions are symmetrical. The model is able to infer the expected time difference on a coincidence-basis according to the input data. Considering the goodness-of-fit, the shapes of the predicted distributions are in very good agreement with the expected Gaussian. A substantial deviation from the Gaussian shape is observed when moving toward the far left and far right source positions. 
A part of the linearity analysis for model \mbox{$(18, 0.1)$} is exemplarily depicted for the position \mbox{$(x_{\text{s}},y_{\text{s}}) = (12,0) \unit{\milli \metre}$} in \cref{fig:example_linReg}. The global $\varepsilon$ performance for each model is shown in \cref{fig:epsilon_perf1}. The estimated $\varepsilon$-parameters of all trained models are within a $3\sigma$-interval in agreement with the theoretical value of $\varepsilon = 1$.

\subsection{CTR Performance}
\label{subsec:CTR_per_res}

\begin{figure}
\centering
\includegraphics[width=\scBroad cm]{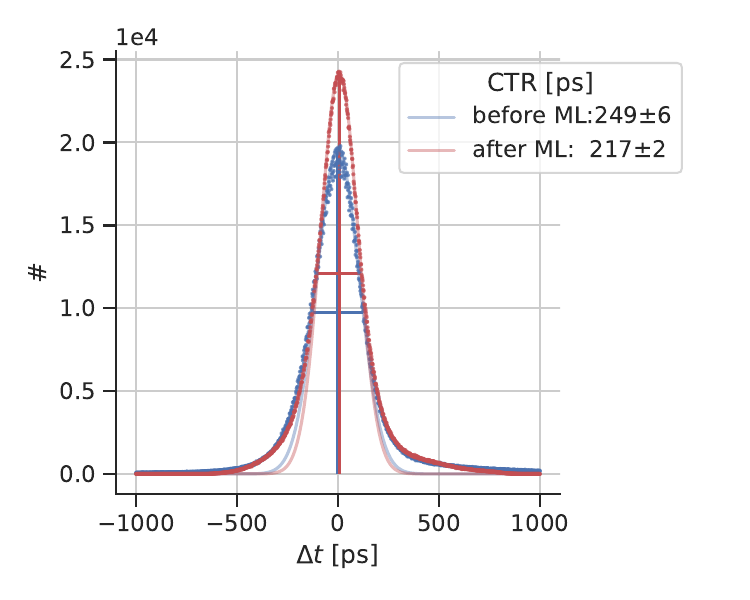}
\caption{Time difference distributions before and after using the proposed machine learning time skew calibration. The model $(18, 0.1)$ was used. No energy windows or restrictions regarding the light distribution are applied.}
\label{fig:ctr_plot}
\end{figure}

The \ac{CTR} performances of the trained models, as well as the performance of applying only the analytical corrections, are listed in \cref{tab:CTR_overview}. As one can see, the best \ac{CTR} was achieved by the model $(18, 0.1)$, which also performed best regarding the \ac{MAE} evaluation. The model improved the \ac{CTR} by about \qty{50}{\pico \second} down to \qty{185 \pm 2}{\pico \second} for an energy window of \qtyrange{450}{550}{\kilo \eV}. Except for the models having a max. depth of $d=12$, all other models yield an improved \ac{CTR} performance when using lower learning rates. A comparison of the time difference distributions before and after using the model $(18, 0.1)$ is depicted in \cref{fig:ctr_plot}. Regarding the shape of the emerging distribution especially coincidences in the tails of the distribution have been recovered to smaller time differences.

\begin{table}
\centering
\caption{\ac{CTR} performance of the trained models based on the performance dataset. The results of applying only the analytical timing calibration is denoted as 'before ML'.}
\label{tab:CTR_overview}
\begin{tabular}{@{}llll@{}}
\toprule
 & \multicolumn{3}{c}{CTR {[}ps{]}} \\ 
\multirow{-2}{*}{Model $(d,lr)$} & all & $[300,700]$ keV & $[450,550]$ keV \\ \cmidrule(r){1-4}
before ML & 249 $\pm$ 6 & 238 $\pm$ 5 & 235 $\pm$ 5 \\ \midrule
(12, 0.1) & 230 $\pm$ 2 & 208 $\pm$ 2 & 197 $\pm$ 2 \\
(12, 0.3) & 224 $\pm$ 2 & 203 $\pm$ 2 & 189 $\pm$ 2 \\
(12, 0.5) & 225 $\pm$ 2 & 204 $\pm$ 1 & 191 $\pm$ 2 \\
(15, 0.1) & 223 $\pm$ 2 & 203 $\pm$ 2 & 190 $\pm$ 2 \\
(15, 0.3) & 225 $\pm$ 2 & 206 $\pm$ 2 & 193 $\pm$ 2 \\
(15, 0.5) & 232 $\pm$ 2 & 211 $\pm$ 2 & 201 $\pm$ 3 \\
\rowcolor[HTML]{C0C0C0} 
(18, 0.1) & 217 $\pm$ 2 & 198 $\pm$ 2 & 185 $\pm$ 2 \\
(18, 0.3) & 226 $\pm$ 2 & 206 $\pm$ 2 & 195 $\pm$ 3 \\
(18, 0.5) & 233 $\pm$ 2 & 214 $\pm$ 2 & 201 $\pm$ 2 \\
(20, 0.1) & 220 $\pm$ 2 & 201 $\pm$ 2 & 188 $\pm$ 2 \\
(20, 0.3) & 225 $\pm$ 2 & 204 $\pm$ 2 & 190 $\pm$ 2 \\
(20, 0.5) & 236 $\pm$ 2 & 216 $\pm$ 1 & 203 $\pm$ 2 \\ \bottomrule
\end{tabular}
\end{table}

\subsection{SHAP Analysis}

\begin{figure}
\centering
\includegraphics[width=\scBroad cm]{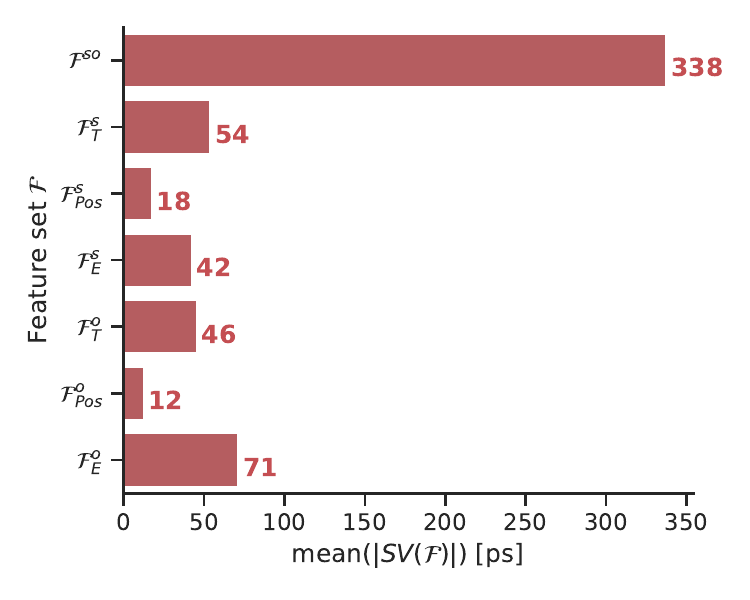}
\caption{The mean absolute SHAP values $\text{mean}(|SV(\mathcal{F})|)$ estimated from a subset of the performance dataset for the different feature sets $\mathcal{F}$ explained in \cref{fig:overview_features}. The strongest contribution comes from the shared feature set $\mathcal{F}^{\text{so}}$, which consists of the difference between the first timestamps $\Delta t_{\text{meas}}$. Furthermore, detector-specific information about the timestamps ($\mathcal{F}_{\text{T}}^{\text{s/o}}$) and energy information ($\mathcal{F}_{\text{E}}^{\text{s/o}}$) are of great importance.}
\label{fig:barplot_shap}
\end{figure}

The model $(18, 0.1)$ was chosen for the analysis using SHAP \cite{Lundberg2017, Lundberg2018, lundberg_local_2020} since it provided the best performance regarding \ac{MAE} and \ac{CTR}. The mean absolute contributions of the different feature sets $\mathcal{F}$ are depicted in \cref{fig:barplot_shap}. The most important feature set is $\mathcal{F}^{\text{so}}$, which consists of the measured time difference $\Delta t_{\text{meas}}$. Besides this substantial contribution, timestamp information $\mathcal{F}_{\text{T}}^{\text{s/o}}$, and energy information $\mathcal{F}_{\text{E}}^{\text{s/o}}$ also seem to be crucial for good model performance. The feature group $\mathcal{F}_{\text{Pos}}^{\text{s}}$ shows a slightly higher contribution compared to $\mathcal{F}_{\text{Pos}}^{\text{o}}$, due to the introduction of \ac{DOI} information. The specific contributions of the planer coordinates, however, differ only marginally for the slab and the one-to-one coupled detector. Furthermore, one observes a similar behavior comparing the feature sets of the slab and the one-to-one coupled detector.

\begin{figure}
\centering
\begin{overpic}[width=\scBroad cm, percent]{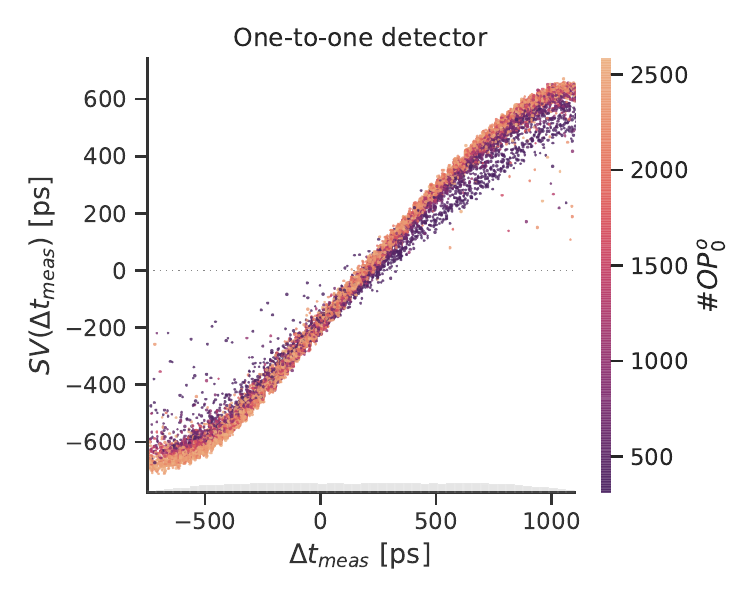}
\put(\subFigX , \subFigY){a)}
\end{overpic}
\begin{overpic}[width=\scBroad cm, percent]{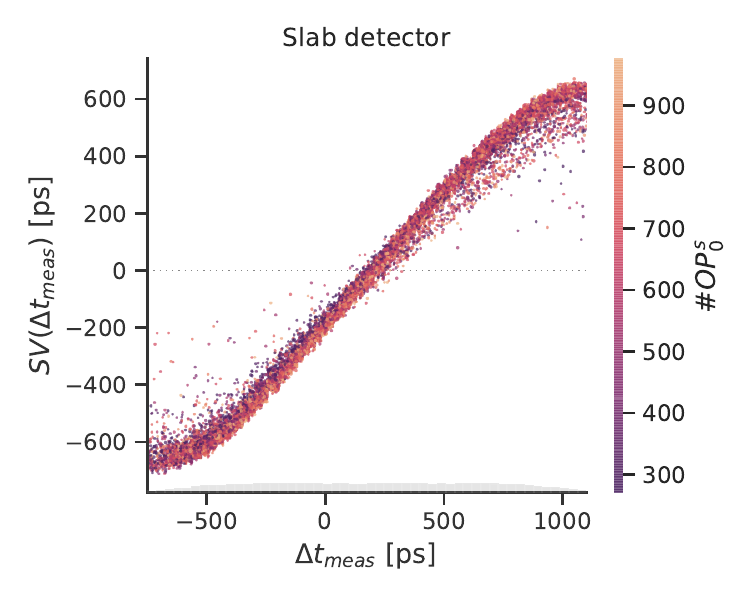}
\put(\subFigX , \subFigY){b)}
\end{overpic}
\caption{Progression of the SHAP values $SV(\Delta t_{\text{meas}})$ in dependence of the feature value $\Delta t_{\text{meas}}$ itself. The number of optical photons detected on the \ac{SiPM} providing the first timestamp ($\#\text{OP}^{\text{s/o}}_0$) is encoded in the color a) for the one-to-one coupled detector and b) for the slab detector.}
\label{fig:timewalk_shap}
\end{figure}

When looking at the progression of the SHAP values $SV(\Delta t_{\text{meas}})$ in dependence on the number of detected optical photons for the \ac{SiPM} providing the first timestamp ($\#\text{OP}_0^{\text{s/o}}$), one observes different developments. \Cref{fig:timewalk_shap}a) shows a clear separation between different SHAP values for a given feature value of $\Delta t_{\text{meas}}$ depending on the number of detected optical photons. This is not observed for the slab detector since, in \cref{fig:timewalk_shap}b), the strong separation regarding the number of detected optical photons is not given.

\section{Discussion}

All models have been trained successfully. The predictions follow a Gaussian function for a large area of the trained data range, as it can be seen in \cref{fig:mae_performance}, \cref{fig:overview_chiq}, as well as in \cref{tab:mae_overview}. When moving to the borders of the presented data, the models' outputs deviate from the expected shape, and the prediction quality decreases. This effect is known for many machine learning algorithms and can be reasoned by the inability to extrapolate to values outside the training range. In future studies, we want to address this issue with different strategies, with using a higher sampling rate at the edges being one of them.\\
Within the central region, where the models show stable behavior, the means of the prediction distributions follow the expected linear relation of \cref{eq:linearity_cond1} (see \cref{fig:example_linReg}). No systematic deviation from the linear relation between offset position $z_{\text{s}}$ and predicted mean time difference $\mu_{\text{s}}$ could be observed, indicating that the trained models are capable of learning the given physical problem. The averaged \mbox{$\varepsilon$-values} are slightly bigger than the expected value of \mbox{$\varepsilon_{\text{theo}} =1$} ($\max(\{\varepsilon_i - \varepsilon_{\text{theo}} \} ) \leq \num[round-mode=places, round-precision=1, exponent-mode=scientific]{0.0038783447208508015}$), which consequently enlarges time differences, and therefore produces an overestimation of the determined \ac{CTR} values, such that a re-scaled resolution might be even better than the here reported one. To compensate for this effect, one could introduce a scaling function $s(\mu_{\text{s}})$ which would correct the slope to the desired value of $\varepsilon=1$ for a given mean time difference. Since the observed effect is insignificant and the estimated slope factors $\varepsilon$ agree for all models with the theoretical value within a $3\sigma$-interval, this procedure is unnecessary for the \ac{GTB} models used in this work.\newline
All trained models can improve the achievable \ac{CTR} values, such that sub-$\qty{200}{\pico \second}$ resolution could be reached for an energy window from \qtyrange{300}{700}{\kilo \eV} (see \cref{tab:CTR_overview}). Minding the shape of the emerging distribution, especially coincidences in the tails of the distribution have been recovered to more minor time differences, indicating that the model can learn physical effects and correct those. This observation underlies the capability of this new approach and shows that the timing resolution can be improved beyond the usage of purely analytical calibrations.\\
We used explainable AI (XAI) techniques to understand on which quantities the models are relying on. The analysis of the SHAP values of the model $(18, 0.1)$ reveals that the reported timestamp difference $\Delta t_{\text{meas}}$ mainly, but also timing and energy information is of great importance. This observation agrees with human intuition, since $\Delta t_{\text{meas}}$ would represent a human's first estimator if one tried to solve the task given to the model. Furthermore, the results clearly indicate that the model is learning timewalk effects for the one-to-one coupled detector (see \cref{fig:timewalk_shap}a), since for a given feature value $\Delta t_{\text{meas}}$, the SHAP value is increased or decreased depending on whether a high or a low number of optical photons has been detected. If a timestamp is affected by timewalk, the exact moment of timestamping is delayed due to low deposited energy. In conclusion, the importance of this timestamp has to be decreased since it would enlarge the reported time difference and worsen the \ac{CTR}. This observation does not occur in the same clearness for the slab detector (see \cref{fig:timewalk_shap}b). However, for the one-to-one coupled detector, the vast majority of information is contained in one channel, whereas, for the semi-monolithic case, the information is spread across multiple channels, making it hard to display the effect in the chosen visualization. There is still an indication that also for the slab detector, timewalk effects are caught by the model since the feature set using energy-related quantities shows a high absolute SHAP value (see \cref{fig:barplot_shap}) and that both tails of the time difference distribution are reduced.

\section{Conclusion \& Outlook}

In this work, we demonstrated a new approach based on the combination of residual physics and machine learning to address real-world physics-based problems. We applied the concept to detector calibration. We hope the work highlights the potential for applications of learning systems along all computing steps of complex acquisition and processing systems and, thus, may inspire future research.\newline
Since the formalism settles on previously linear corrected timestamps \cite{naunheim_analysis_2022}, it can be seen as a first approach towards residual physics in timing calibration. All models could be trained successfully and are in a $3\sigma$-agreement with the underlying physical relation. The first results indicate that this new calibration strategy has provoked a strong improvement in the achievable \ac{CTR} reaching from \qty{238}{\pico \second} down to \qty{198}{\pico \second} for an energy window of \qtyrange{300}{700}{\kilo \eV}, and from \qty{235}{\pico \second} even down to \qty{185}{\pico \second} for a smaller energy window of \qtyrange{450}{550}{\kilo \eV}. The SHAP analysis offers a strong indication, that the proposed technique has the capability to build physics-informed models.\\
All results are based on experimentally acquired data from two clinically relevant detector arrays. This work and the corresponding promising first results represent a proof-of-concept for future time skew calibration techniques relying on \ac{AI}. Nevertheless, several studies have to be performed before an application to a complete \ac{PET} system is possible. The presented technique is currently implemented for a pair of detectors utilizing digital \acp{SiPM}. Research towards systems of multiple detectors will be addressed in future works. Besides this, we want to explore the performance of the concept in different environmental settings (e.g., higher measurement temperatures, different readout), potentially enlarging the learning system's importance. Furthermore, the reduction and study of the influence of the needed data acquisition time and the bias effects towards the edges of the training data is mandatory for a possible usage in a clinical scanner. A possible method to address this point would be an artificial enlargement of the available training data found on only a few measured data points. However, we expect the measurement time to increase weaker than linearly with the number of detectors since one source position can be used for many detectors.

\section*{Acknowledgement}
The work was funded by the German Federal Ministry of Education and Research under contract number 13GW0621B within the funding program ‘Recognizing and Treating Psychological and Neurological Illnesses - Potentials of Medical Technology for a Higher Quality of Life’(‘Psychische und neurologische Erkrankungen erkennen und behandeln - Potenziale der Medizintechnik für eine höhere Lebensqualität nutzen’).

\bibsetup
\printbibliography

\end{document}